\documentclass[conference]{IEEEtran}
\IEEEoverridecommandlockouts
\usepackage{cite}
\usepackage{graphicx}
\usepackage{amsmath,amssymb,amsfonts}
\usepackage{algorithmic}
\usepackage{graphicx}
\usepackage{textcomp}
\usepackage{xcolor}
\usepackage{amsmath,amssymb,amsfonts}
\usepackage{algorithmic}
\usepackage{graphicx}
\usepackage{textcomp}
\usepackage{xcolor}
\usepackage{multirow}
\usepackage{caption}
\usepackage{soul}
\usepackage{afterpage}
\usepackage{array}
\usepackage{threeparttable}
\usepackage{subcaption}
\def\BibTeX{{\rm B\kern-.05em{\sc i\kern-.025em b}\kern-.08em
    T\kern-.1667em\lower.7ex\hbox{E}\kern-.125emX}}

\makeatletter
 \let\old@ps@headings\ps@headings
 \let\old@ps@IEEEtitlepagestyle\ps@IEEEtitlepagestyle
 \def\confheader#1{%
 \def\ps@headings{%
 \old@ps@headings%
 \def\@oddhead{\strut\hfill#1\hfill\strut}%
 \def\@evenhead{\strut\hfill#1\hfill\strut}%
 }%
 \def\ps@IEEEtitlepagestyle{%
 \old@ps@IEEEtitlepagestyle%
 \def\@oddhead{\strut\hfill#1\hfill\strut}%
 \def\@evenhead{\strut\hfill#1\hfill\strut}%
 }%
 \ps@headings%
 }
\makeatother

\begin{document}

\title{Radar Guided Dynamic Visual Attention for Resource-Efficient RGB Object Detection\\}

\confheader{%
Accepted in International Joint Conference on Neural Networks (IJCNN) 2022}

\author{\IEEEauthorblockN{Hemant Kumawat}
\IEEEauthorblockA{\textit{School of Electrical and Computer Engineering} \\
\textit{Georgia Institute of Technology}\\
hkumawat6@gatech.edu}
\and
\IEEEauthorblockN{Saibal Mukhopadhyay}
\IEEEauthorblockA{\textit{School of Electrical and Computer Engineering} \\
\textit{Georgia Institute of Technology}\\
saibal.mukhopadhyay@ece.gatech.edu}}

\maketitle

\begin{abstract}

 An autonomous system's perception engine must provide an accurate understanding of the environment for it to make decisions. 
 Deep learning based object detection networks experience degradation in the performance and robustness for small and far away objects due to a reduction in object's feature map as we move to higher layers of the network.
 In this work, we propose a novel radar-guided spatial attention for RGB images to improve the perception quality of autonomous vehicles operating in a dynamic environment. In particular, our method improves the perception of small and long range objects, which are often not detected by the object detectors in RGB mode. The proposed method consists of two RGB object detectors, namely the Primary detector and a lightweight Secondary detector. The primary detector takes a full RGB image and generates primary detections. Next, the radar proposal framework creates regions of interest {(ROIs)}  for object proposals by projecting the radar point cloud onto the 2D RGB image. These ROIs are cropped and fed to the secondary detector to generate secondary detections which are then fused with the primary detections via {non-maximum suppression}. This method helps in recovering the small objects by preserving the object's spatial features through an increase in their receptive field. We evaluate our fusion method  on the challenging nuScenes dataset and show that our fusion method with SSD-lite as primary and secondary detector improves the baseline primary yolov3 detector's recall by 14\% while requiring three times fewer computational resources. 



\end{abstract}


\section{Introduction}
Object detection is one of the most challenging tasks in autonomous systems. 
Deep convolutional architectures such as R-CNN \cite{rcnn}, Fast R-CNN\cite{DBLP:journals/corr/Girshick15}, and Faster RCNN\cite{NIPS2015_5638} provide very high accurate object detection results. However, due to slow inference time and high memory requirements\cite{p_rccn}, single stage detectors like SSD \cite{DBLP:journals/corr/LiuAESR15} and YOLO\cite{yv3} are used for faster detection often at a slightly lower accuracy. However, single stage RGB detectors suffer from degradation in performance on small object detection task. Object detection becomes very challenging as the object height relative to the image size decreases\cite{tile}. In addition to this, these object detectors perform very poorly when they are exposed to novel environments or adverse lighting and weather conditions\cite{rain}\cite{fog}
. These perception failures could be very critical to the autonomous system's decision making module and may cause catastrophic failures.


To overcome the limitations of RGB camera based perception, modern day autonomous systems are often equipped with multiple perception modalities like cameras, radars, and LiDARs\cite{DBLP:journals/corr/abs-1912-04838}\cite{Geiger2013IJRR}
. Using multiple sensors allows the perception engine of the autonomous systems to exploit complementary features provided by different modalities. Automotive radars have been widely used in vehicles for Advanced Driving Assistance Systems (ADAS)\cite{ziebinski2017review} as they can detect long range objects and are highly robust to adverse weather and lighting conditions.  Hence, a radar could provide valuable information of long range objects that are missed by the RGB detectors.

Although radar could detect objects at long range accurately, processing radar point cloud is a very challenging problem due to the unstructured nature of the data. The radar's pointcloud is very sparse with inconsistent point density for objects without any labels. Classical methods\cite{c1}\cite{c2}\cite{c3}
for fusing camera and radar includes kinetic model based tracking and filter based association algorithms. However, noisy and sparse 3D radar data makes the association problem challenging. Often this process is handcrafted with some heuristic rules to make it work with 2D RGB images. An emergent solution is to use radar point based feature extraction methods. These methods are usually designed for dense LiDAR pointclouds such as PointNet\cite{qi2017pointnet} and hence do not have a good performance on sparse and noisy radar data. Other methods include creating a depth map using raw radar data and then fusing the depth map with RGB images\cite{LeeWeiYuLee2021}\cite{n1}
. However, many of these methods require two stage detectors\cite{bira} making them computationally expensive for resource constrained systems. Even though methods with radar guided feature extraction have better results on small object detection, they are far from real time deployment on robotics systems due to their high computational needs.  

This paper introduces a novel fusion method for RGB camera and radar to detect small and far away objects in RGB images with low complexity and memory footprint. The proposed method is designed with a lightweight single stage detector that fuses radar point cloud with RGB images to generate ROIs for object candidates. Radar point cloud contains range, azimuthal and velocity information of objects with high confidence. Each point in the radar pointcloud when projected onto the image plane gives an approximate location of the object in the image. We use radar points to generate regions of interests with object candidates. These ROIs are fed to the lightweight detector to generate object detections. This increases the relative size of the object with respect to the ROI image and thereby, reducing the information loss of object features while convolution and pooling operations. The final step includes a fusion of these detections with detections made by passing the full image to another detector using NMS. This way we recover any large objects missed or not fully detected due to cropping of the image. We refer to this network as 'Primary detector' and our lightweight detector as 'Secondary detector'. We evaluate our network on the nuScenes dataset \cite{caesar2020nuscenes} which provides synchronized data from a full autonomous sensor suite including multiple radars and RGB cameras. Our experiments show that the proposed method shows  14\% increase in recall for fusion system with SSDlite\cite{ssdlite} as primary and  secondary detector than baseline yolov3 detector while only requiring 22.3 GFLOPs against 66.4 GFLOPs of the primary detector.

\section{Fusion Framework}

\subsection{Motivation} 
\begin{figure}
\centering
\includegraphics[width=\linewidth]{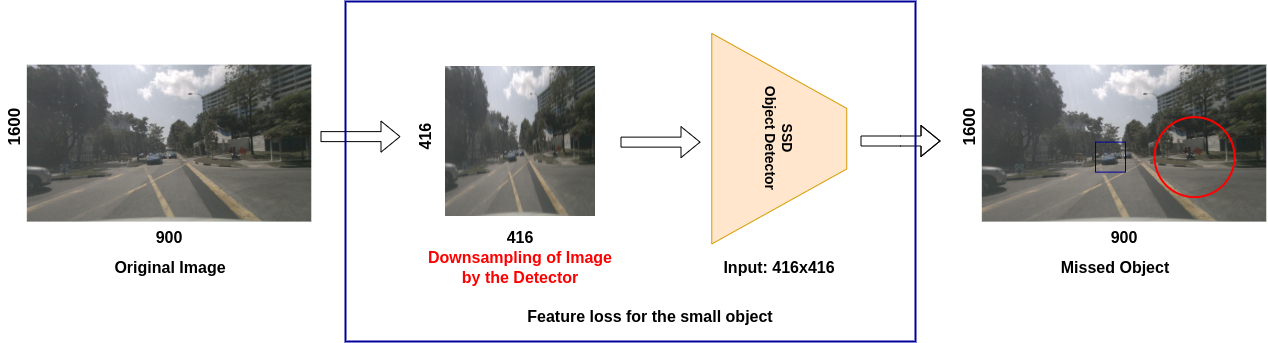}
\caption{ Feature loss for small object\label{fig:primary}}
\end{figure}

%
Our understanding of how well the current object detectors perform on detection task is largely measured by observing average recall and precision on benchmark datasets. However, there is a big gap between the detectors' performance on small and large objects which is not reflected in average performance numbers. This is true for any type of object detector and this gap increases for detectors with a small computational footprint\cite{liu}\cite{DBLP:journals/corr/HowardZCKWWAA17}. The small objects in the images lack the distinctive features and when passed through the convolutional backbone of the network, background and nearby pixel features begin to dominate in higher layers as shown in figure \ref{fig:primary}. Low level features generated using lower layers do not have enough semantic information to help in prediction which makes this problem very challenging. 


One of the ways to increase the receptive field of the small object is to scale the image to a large size and then pass it through the detector. This reduces the feature corruption for the object with background and nearby pixels, thereby helping in better performance on small objects. Authors in \cite{Hu} use two object detectors for the face detection task with a second detector processing the upscaled image to detect small objects. However, increasing the input size for detectors increases the computational overhead of the system significantly making it impractical for systems with limited computational power. Table \ref{tab:yolo_gflops} shows the effect of increasing the input scale for yolov3-spp on recall and GFLOPs. We observe an increase in recall of the detected objects as the input size is increased from 256 to 1080, however it increases the computational load by almost 17 times. Note that the relation between the number of GFLOPs and input size is not linear and it increases { non-linearly} as the input size is scaled.   

The receptive field for the small object could also be increased by cropping a small region around the object and passing it through the detector. This increases the normalized area of the object pixels relative to the cropped image thereby reducing the feature loss to the background and nearby pixels. However, single detectors  are not designed to create proposals for the object regions. In absence of proposal generators for object regions\cite{tile}, many works try to divide the image into multiple tiles and pass them through the detector. However, this design may require a large number of regions created randomly in the image to detect all the small objects. This may severely affect the  computational efficiency of the network leading to a higher inference time. 

Radar's power of detecting objects at long range could be very helpful in guiding where the object is in the image. Radar detection when mapped to the image gives an approximate location of the object which could be used to create proposals for the object regions. Many of the automotive radars have very sparse data with less than 64 points making them excellent for our purpose. There have been prior works where the radar data is used to guide the proposal generation stage in two stage detectors\cite{chadwick}\cite{yang}. Yadav et. al\cite{bira} created  a network BIRANet which fused radar points with a feature extractor network to guide anchors generation for Faster R-CNN. However, the anchors generated using this method do not increase the receptive field and only act as a proposal generator. Moreover, they require high energy resources and have a longer inference time due to the use of two stage detectors.

\begin{table}[t!]
    \centering
    {
	\captionsetup{justification=centering}
	\caption{Comparison between performance and computation overhead for yolov3 with different input size}
	\setlength{\belowcaptionskip}{-10pt}
	\begin{tabular}{c c c }
		\hline
		\textbf{Image size} & \textbf{Recall} & \textbf{GFLOPs} \\
		\hline
		256 & 0.29 & 25.1  \\
		 416& 0.42 & 66.4  \\
		 512& 0.45 & 100.5  \\ 
		 640& 0.49& 157.1  \\

		1080 & 0.54 & 447.4  \\
		1920 & 0.49 & 1413.9  \\
		 \hline
	\end{tabular}
	\label{tab:yolo_gflops}
	}
\end{table}





\subsection{Proposed Architecture}
\begin{figure*}
\centering
\includegraphics[width=\linewidth]{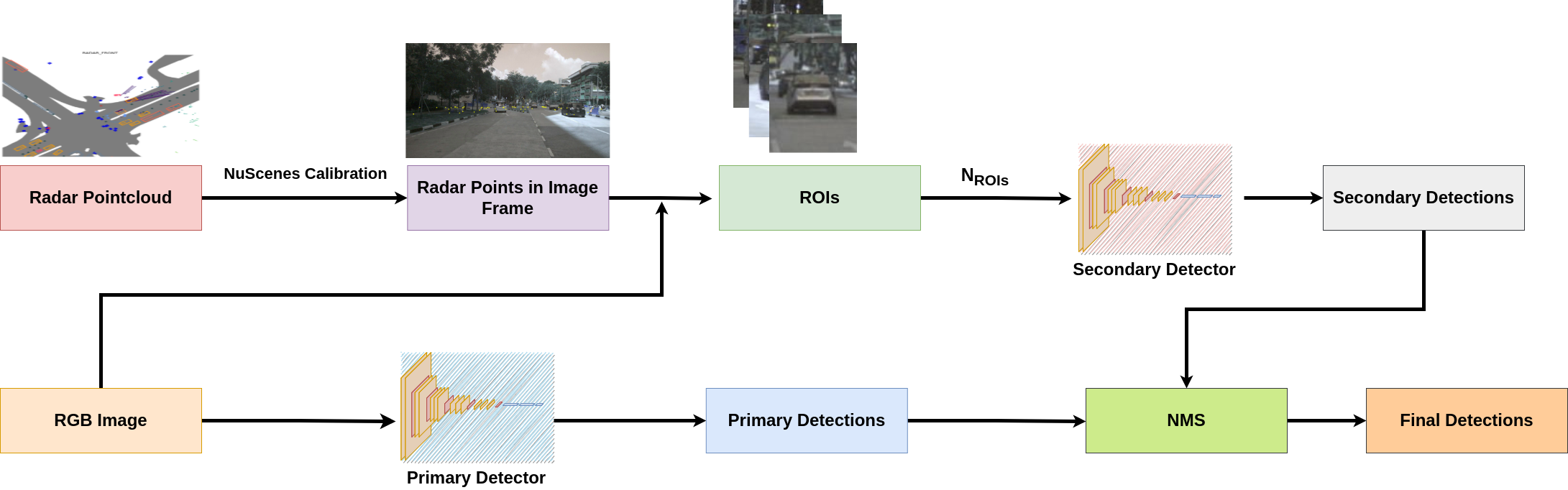}
\caption{ Algorithmic pipeline of Proposed Fusion System\label{fig:1}}
\end{figure*}

In this study, we aim to use the radar points to generate object proposals in the image and use lightweight detectors to detect objects in the proposal regions. Our proposed fusion network is shown in Figure \ref{fig:1}. The proposed network consists of two detectors: a primary detector and a secondary lightweight detector. The primary detector takes the RGB images and generates detection for the full image. We refer to these detections as primary detections. Next, the radar points are projected to the RGB image frame to generate object proposals. These object proposals regions of a predefined size are cropped from the image and each of them is sent to the secondary detector to generate secondary detections. The secondary detections are then merged with the primary detections to generate final detections. The use of radar data ensures that we do not create redundant proposals for detection with no objects. Unlike tiling methods, this method generates proposals in a dynamic fashion. The overall computational cost of the system will depend on the choice of lightweight detectors and the number of radar points. For automotive radars like Aptiv ESR 2.5\cite{autonomoustuff} that  detect less than 64 objects at any timestamp, the maximum computational overload will be 64 times the GFLOPs of the lightweight detector. For a suitable choice of lightweight detector, this could be very well managed by a resource efficient system. 


\subsubsection{Radar Object Proposals}

\begin{figure}
     \centering
     \begin{subfigure}[b]{0.24\textwidth}
         \centering
         \includegraphics[width=\linewidth]{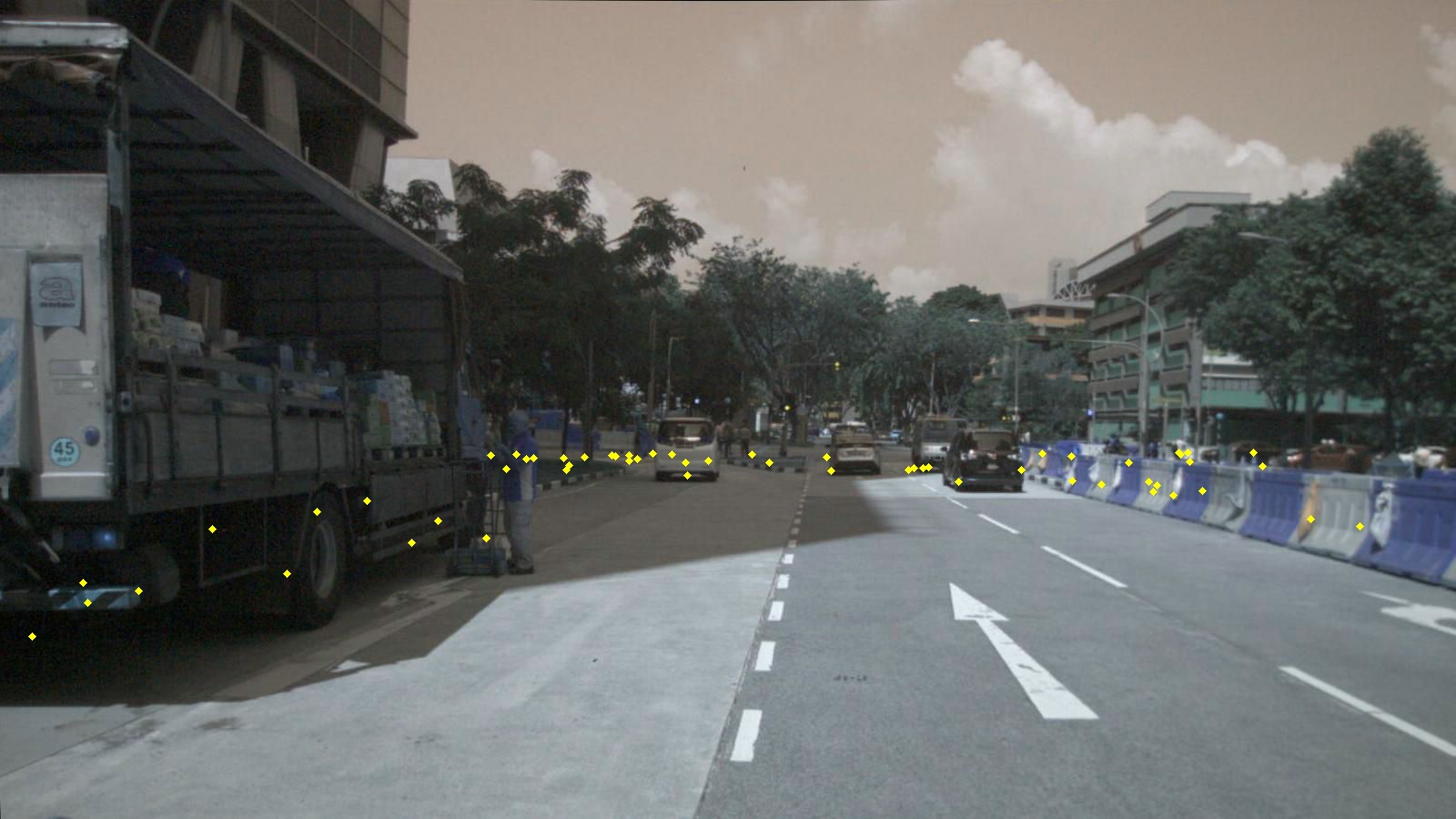}
        \caption{}
         \label{fig:map}
     \end{subfigure}
    \hfill
     \begin{subfigure}[b]{0.24\textwidth}
         \centering
         \includegraphics[width=\linewidth]{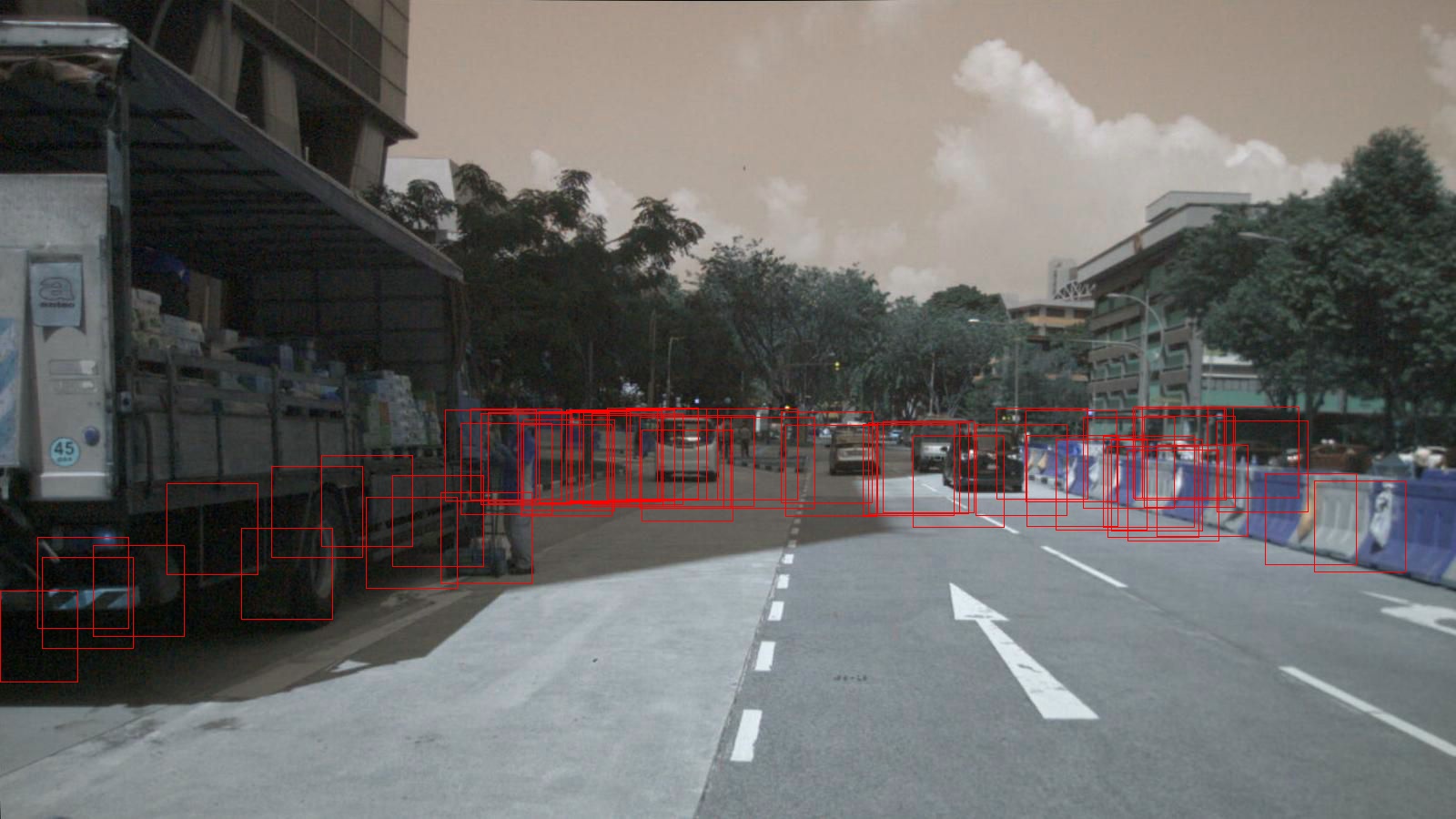}
         \caption{}
           \label{fig:box}
     \end{subfigure}
     \hfill 
 
        \caption{a.) Yellow points in the image show radar points mapped to the camera image, b.) Radar based proposals generated for secondary detector }
        \label{fig:three graphs}
\end{figure}

Detected objects by automotive radars are reported as 3D points in bird's eye view. In addition to position and depth, radars also report the radial velocity of the moving objects. For our method, we parameterize the radar detection as D = (x, y, z) and treat every detection as stand-alone detection. As a pre-processing step, we perform spatial alignment of radar data into the camera's coordinate system. The nuScenes dataset provides us with the necessary calibration tools for mapping radar points to the egocentric and camera frame. When mapped onto the camera frame, radar detections point to the detected objects in the frame. Even though not all the objects are detected by radar, detections include most of the objects in long range detected with high confidence. Each radar point in camera frame is parametrized as P = (cx,cy). We treat this as the approximate location of an object and draw a 2D anchor of predefined size for every point with P as the center. Figure \ref{fig:map} shows radar points in yellow mapped to the camera frame while ROIs generated using radar points are shown in red boxes in figure \ref{fig:box}.

\subsubsection{Object Detection}

The secondary detector is responsible for detecting the objects in the ROIs generated by radar object proposal network. For each radar point, the region bounded by the anchor box is cropped from the image and sent to a lightweight detector. For each image, we generate n ROIs for secondary detection, where n is the total number of radar points in the given frame. All the detections for n ROIs are aggregated with detections by the primary detector generated by passing the full image. Finally, we perform Non-maximum Suppression(NMS) on the aggregated detections to filter out any double detections.

\begin{figure}
     \centering
     \begin{subfigure}[b]{0.24\textwidth}
         \centering
         \includegraphics[width=\linewidth]{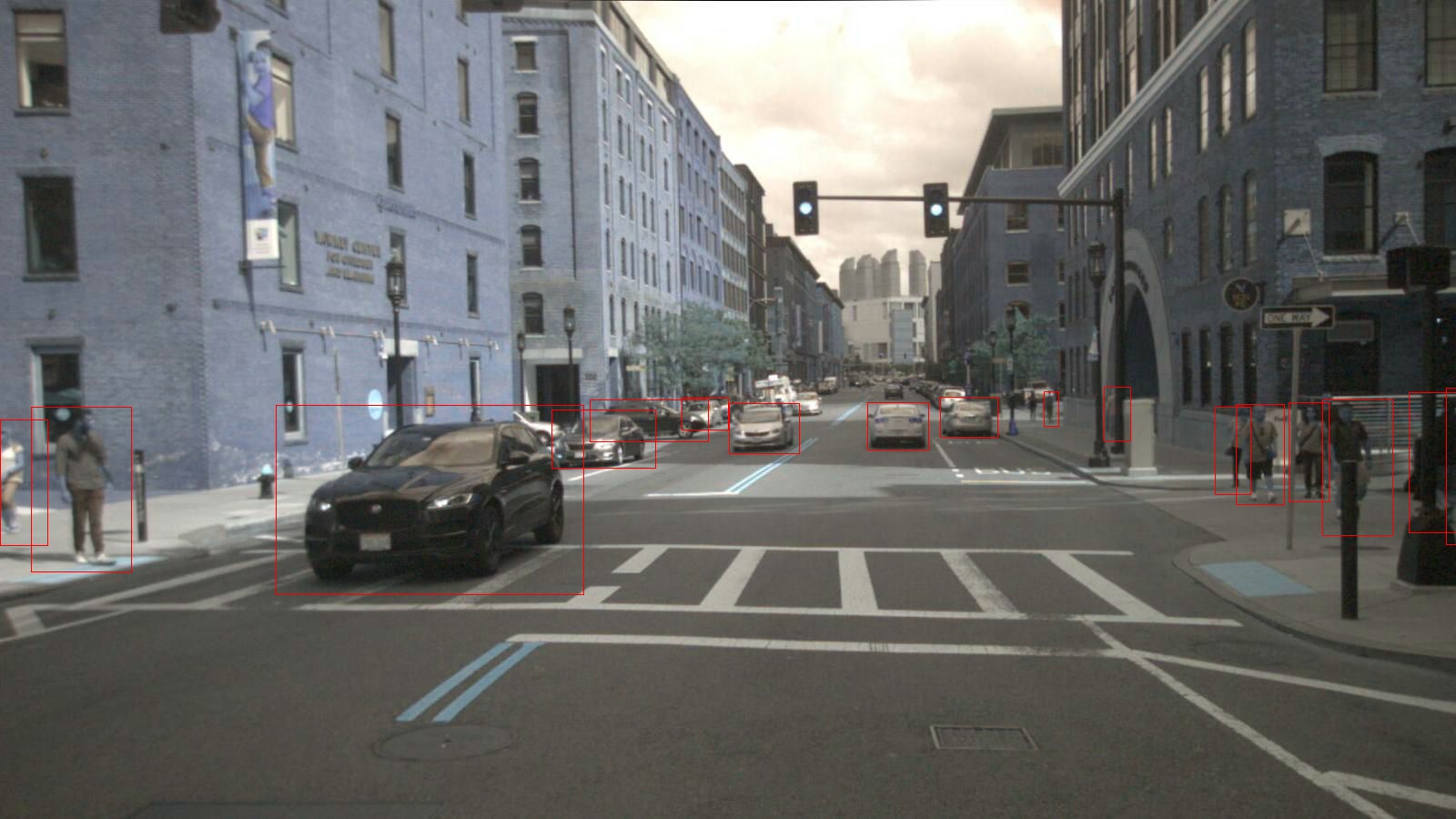}
     \end{subfigure}
     \begin{subfigure}[b]{0.24\textwidth}
         \centering
         \includegraphics[width=\linewidth]{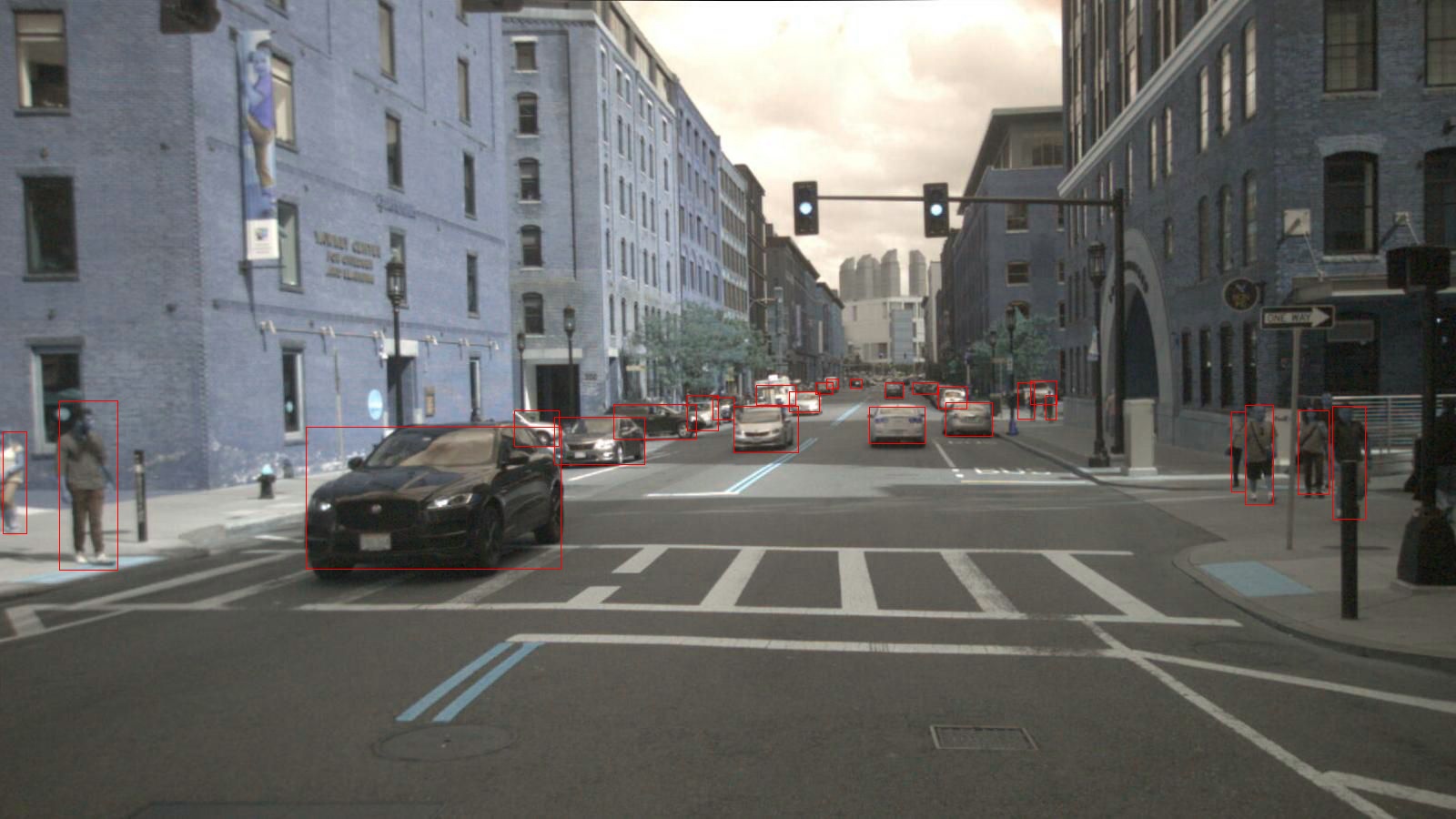}
     \end{subfigure}
     \hfill 
     \begin{subfigure}[b]{0.24\textwidth}
         \centering
         \includegraphics[width=\linewidth]{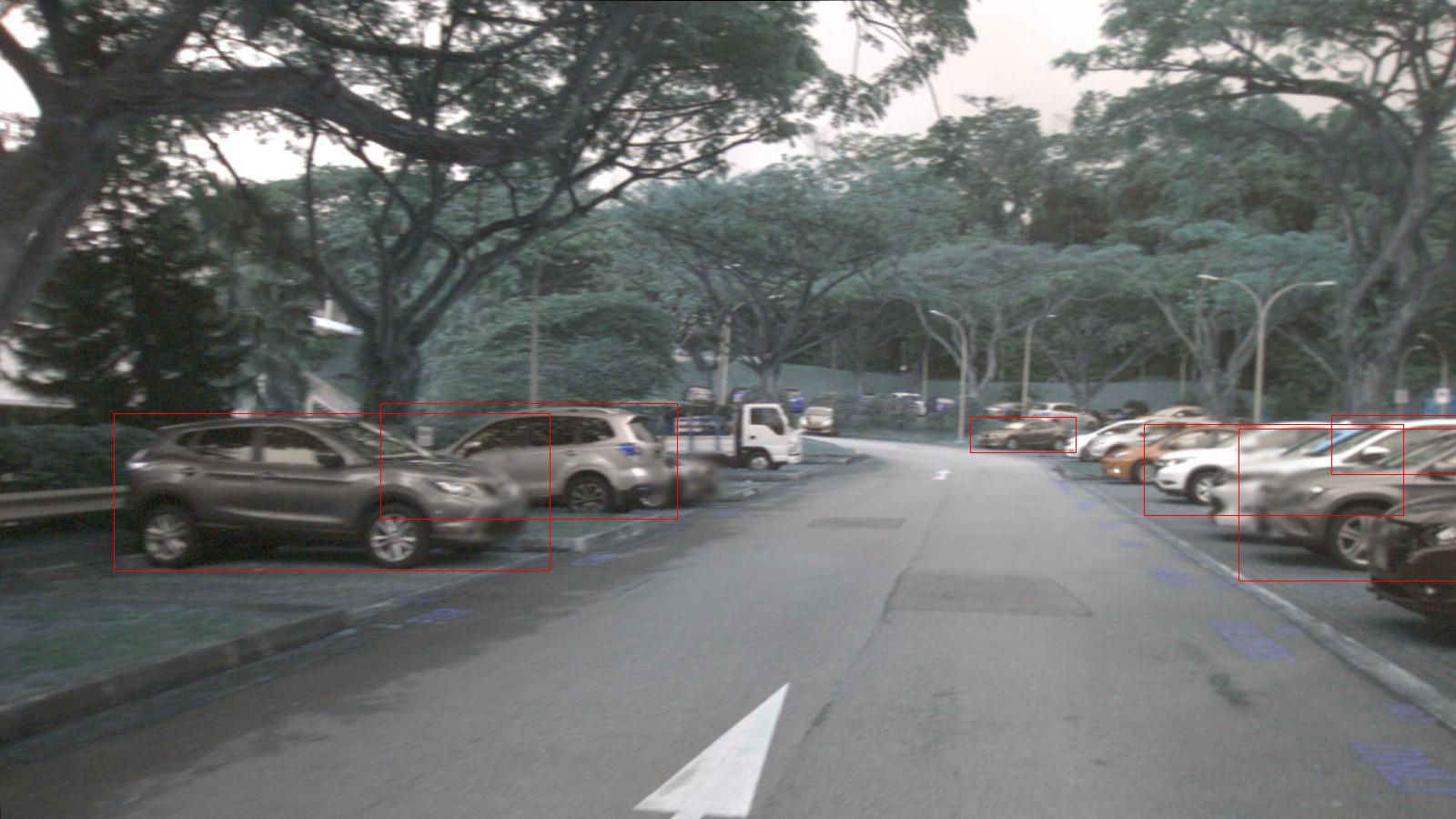}
         \caption{2D annotations in nuScenes}
         \label{fig:five over x}
     \end{subfigure}
     \begin{subfigure}[b]{0.24\textwidth}
         \centering
         \includegraphics[width=\linewidth]{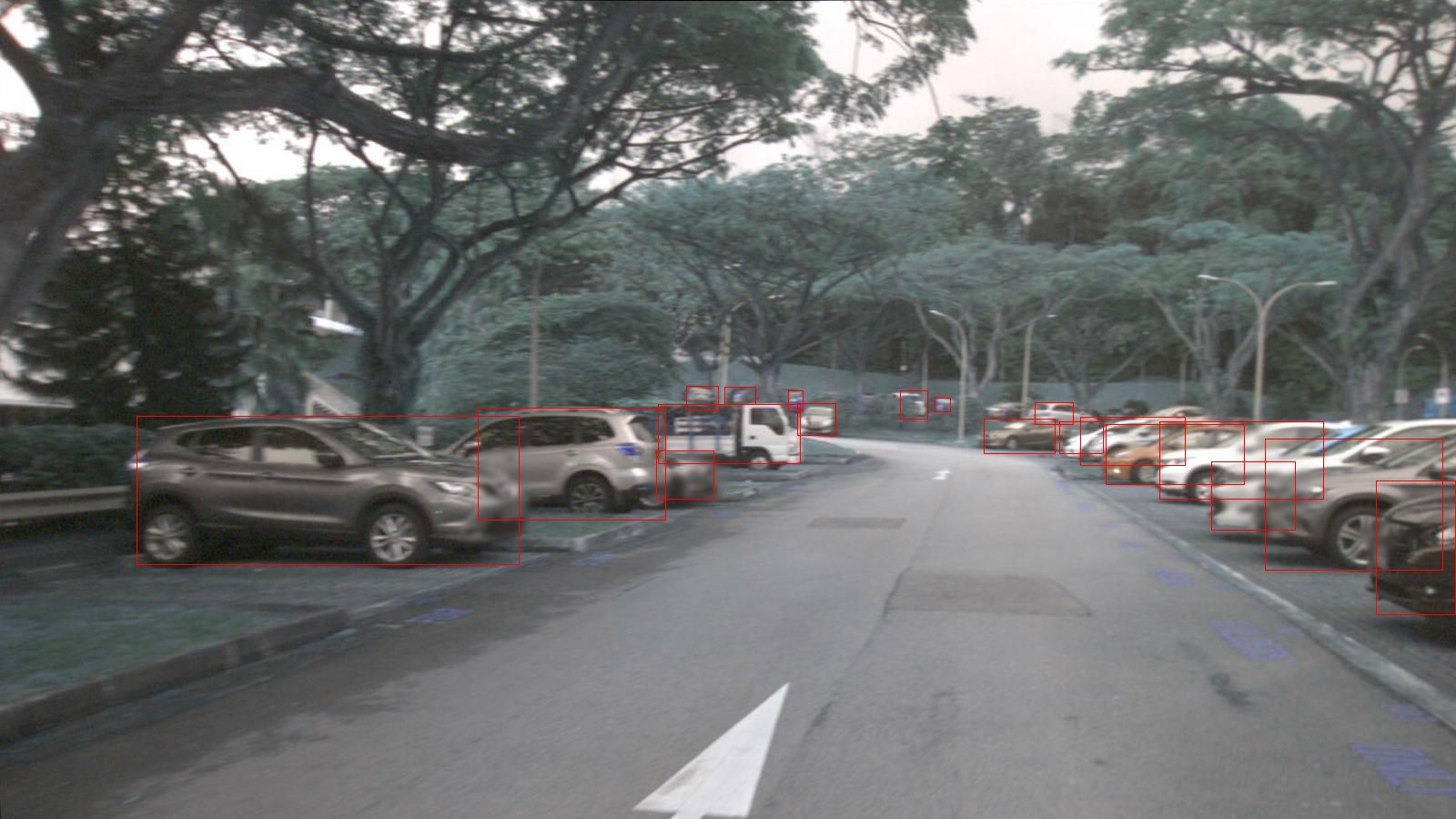}
         \caption{2D annotations labelled by us}
         \label{fig:five over x}
     \end{subfigure}
        \caption{Comparison of Nuscenes 2D annotations and our labelled annotations for images from front camera}
        \label{fig:ann}
\end{figure}


\section{Experimental Results}

\subsection{Dataset}
We test our proposed fusion method on the validation set of the nuScenes \cite{caesar2020nuscenes} dataset that has synchronized data collected from an autonomous vehicle sensor suite of 6 cameras, 5 radars, and 1 lidar, all with a full 360-degree field of view. Since nuScenes has 3D annotations, we first convert these 3D annotations to 2D annotations and merge all relevant classes into 6 classes: car, truck, bus, pedestrian, bicycle, and motorcycle. For our evaluation purposes, we discard highly occluded objects from ground truths annotated in 3D. As shown in figure \ref{fig:ann}, we observe that many of the far-away objects in the scenes did not have any annotations. This is due to the fact that objects that are not covered with at least one lidar or radar point are discarded by nuScenes even though they were captured by RGB cameras. To overcome this, we annotate a small sample of images manually from the \textit{mini-val} split of nuScenes. Our dataset contains 396 images and 3777 annotations with annotated classes of: ’cars,’ ’pedestrian’, ’truck’, ’trailer’, ’bus’, ’cycle’, and ’motorcycle’. The predicted detections are compared against our manually annotated dataset since it has more labels as compared to nuScenes 2D annotations.


\begin{table}
    \centering
    {
	\captionsetup{justification=centering}
	\caption{FLOPs of Object Detectors}
	\setlength{\belowcaptionskip}{-10pt}
	\begin{tabular}{c c c }
		\hline
		\textbf{Detector} & \textbf{Image size} & \textbf{GFLOPs} \\
		\hline
		yolov3-spp & 416 & 66.4 \\
		 & 640 & 157   \\
		 & 1080 & 447   \\ 
		 & 1900 & 1384.6  \\
		\hline
		tiny-yolov3 & 200 & 1.3  \\
		& 300 & 2.9  \\
		& 400 & 5.2 \\
		& 600 & 14.1  \\
		\hline
		SSDlite & 200 & 0.20    \\
		 & 300 & 0.43 \\
		 & 400 & 0.74\\
		 \hline
	\end{tabular}
	\label{tab:all_latency_table}
	}
\end{table}

\subsection{Object Detection network}
We use yolov3-spp, tiny-yolov3 and SSDlite as our primary and secondary object detector for our experiments. It is important to note here that, this approach is applicable to any combination of primary and lightweight secondary detectors to obtain significant detection improvement at low computational power. All the object detectors were trained on the COCO dataset \cite{COCO1} and detection with classes belonging to 'Car', 'Person', 'Bicycle', 'Motorcycle', 'Bus' and 'Truck' considered for evaluation. We compare our fusion system with the detections from the primary detectors.  We used PyTorch to implement our system and all the experiments were  conducted  on  a  setup  with  two  Nvidia  1080Ti GPUs.

\subsection{Metrics For Evaluation}

\begin{table}[t]
    \centering
    {
    \begin{threeparttable}[b]
    \caption{Value Table.}
    \label{tab:value_table}
	\begin{tabular}{ |p{4cm}|p{3.5cm}|}
		\hline
		\textbf{Name} &  \textbf{Value}\\
		\hline
		$\alpha^{*}$ & 20mJ/frame      \\
		$\beta^{1}$  & 0.92J/frame    \\
		$\gamma^{*}$  & 3.9mJ/Mb     \\
		$fs$ & 34.5Mb \\
		$fps$ & 20 \\
		$EE_{hw}$ & 3.08 TOPS/W \\
		\hline
	\end{tabular}
	\begin{tablenotes}
     \item[*] Values adapted from \cite{hemant}
     \item[1] Values calculated from Delphi ESR-2.5 datasheet \cite{autonomoustuff} with max frame-rate=13fps,
   \end{tablenotes}
  \end{threeparttable}
  }
\end{table}
To compare the detection ability of different object detection systems, we use Recall and the number of False Negatives as our metric. Recall measures the number of correctly detected objects over the number of objects in the ground truth. Following is the equation for Recall:

\resizebox{1\linewidth}{!}{
  \begin{minipage}{\linewidth}
\begin{align}
    &Recall = \frac{True Positives}{TruePositives + False Negatives} \nonumber \\
    \label{eqn:metrics}
\end{align}
\vfill
\end{minipage}
}

For comparison between different detectors based on the computation overhead, we compute the number of floating operations executed per second. Table \ref{tab:all_latency_table} shows the comparison of computational overhead in GFLOPs for yolov3-spp, tiny-yolo and SSDlite. We also compute the total energy consumed per frame for different object detectors. This includes energy consumption for sensor activations, data transfer of captured frames and computational expenses due to object detectors. Based on \cite{closedloop}, equations for resource consumption are described below:

\resizebox{1\linewidth}{!}{
  \begin{minipage}{\linewidth}
\begin{align}
    &Energy = \alpha + \beta  + \gamma* f_s + \frac{(CL_{pd} + CL_{sd}*N^{ROIs}).fps}{EE_{hw}} 
    \label{eqn:energyeqn}
\end{align}
\vfill
\end{minipage}
}
Here, $\alpha$, $\beta$ are the per frame sensor's data capture energy for RGB cameras and radar respectively while $\gamma$ is the energy spent on transferring captured frames to the perception engine of the system. $CL_{pd}$ and $CL_{sd}$ are the computational overhead of the primary and secondary object detector respectively.   $N^{ROIs}$ represent the number of ROIs created by the radar proposal network per frame that are processed by the secondary detector. $f_s$ is the combined frame size memory for the radar pointcloud and camera image. We fix the nuScenes frame rate($fps$) to 20 for our experiments.  UNPU \cite{unpu} is used as the DNN accelerator to estimate the computational energy required to run the object detector and $EE_{hw}$ denotes the energy efficiency of our DNN accelerator.  For our calculations, we assume that our system has an automotive radar of 77GHz frequency and an RGB monocular camera that are always turned on. Table \ref{tab:value_table} shows the values of the variables in equation \ref{eqn:energyeqn} used for our evaluation. 

\subsection{Hyperparameter Study}

\begin{figure}
    \centering
    \includegraphics[width=\linewidth]{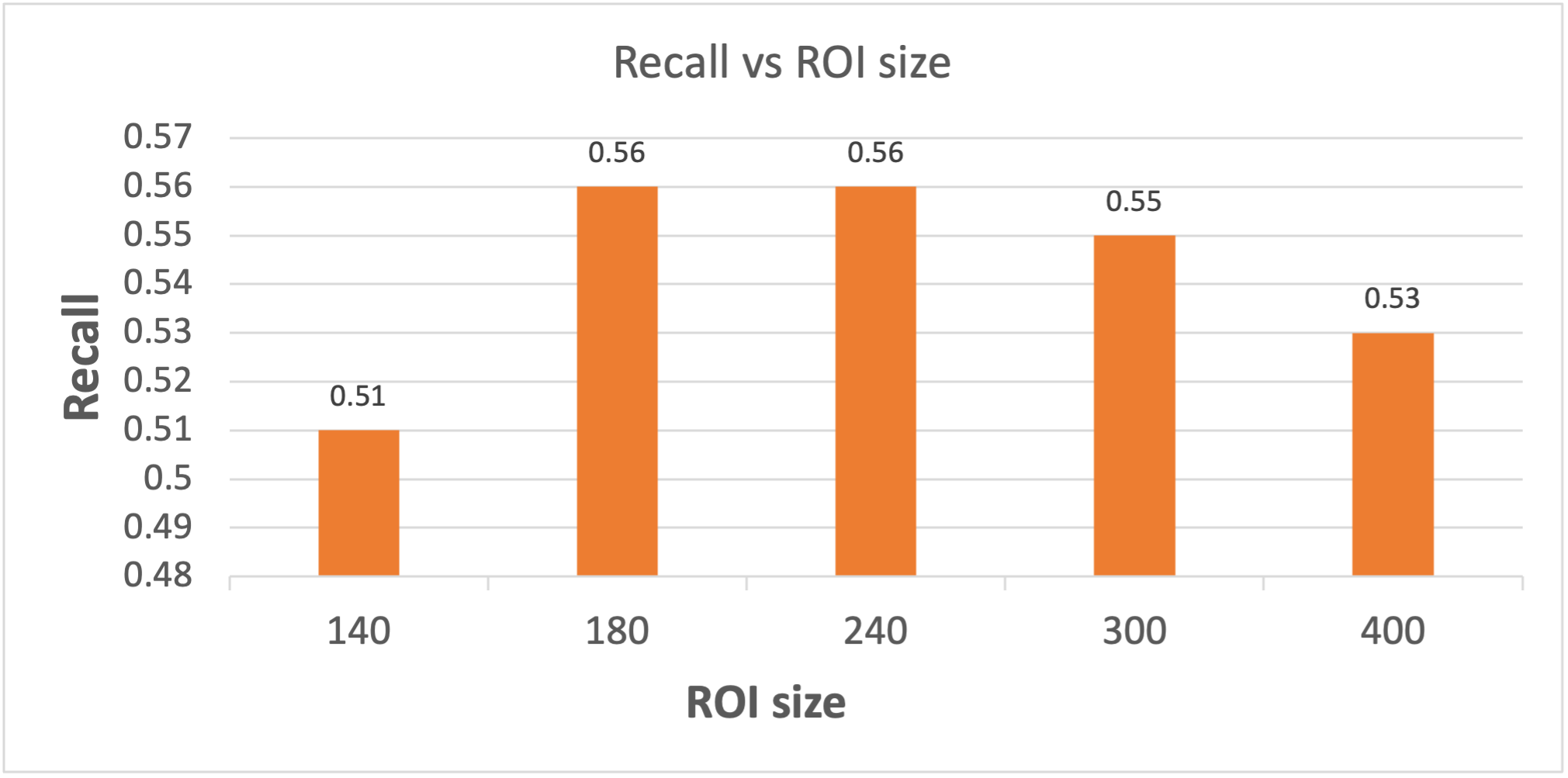}
    \caption{{ROIs size vs Recall comparison with yolov3-spp as primary detector and tiny-yolov3 as secondary detector}}
    \label{fig:roi_size}
\end{figure}
\begin{figure}
    \centering
    \includegraphics[width=0.5\linewidth]{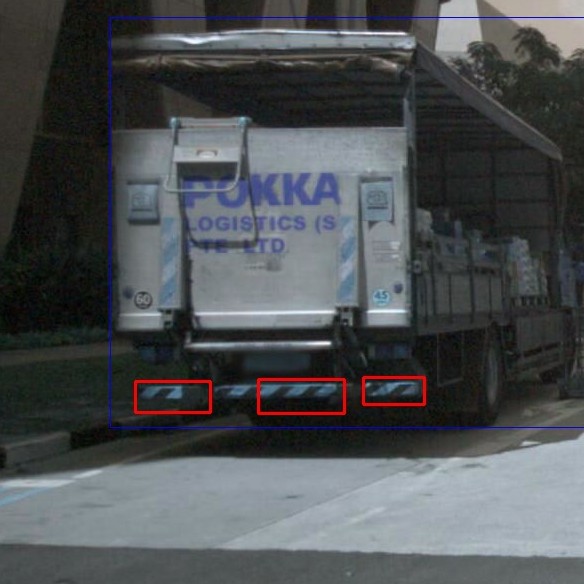}
    \caption{{Multiple detection for small ROI sizes not removed after NMS}}
    \label{fig:doubledet}
\end{figure}

Our proposed algorithm's performance depends on the choice of ROI size of the proposal and  input image size for the lightweight detector since these determine the effective receptive field of the objects in ROIs.  A large ROI proposal will decrease the effective stimuli of the object thereby leading to a missed detection while a small ROI may crop some pixels of the object necessary for the detection . The receptive field of the object increases as the input size of the detector is increased, however it leads to an increase in the computational overhead of the network. Hence, it is important to find appropriate ROI and input image size for the secondary detector to balance the performance and memory footprint of the network.

\subsubsection{Effect of ROI size on recall}
We compared the performance of our fusion method for different input image sizes of the secondary detector and ROI size for secondary detections. Figure \ref{fig:roi_size} shows the recall for different ROIs sizes. We performed the experiments with yolov3-spp as the  primary detector and tiny-yolov3 as the secondary detector with input sizes of 416 and 300 respectively and a low IoU of 0.1 to recover as many false negatives as possible. It is to be noted that changing the ROI size has no effect on overall energy consumption as the energy consumption only depends on the input image sizes of the detectors. We find the highest recall for ROI sizes of 180 and 240 and the performance degrades if the ROI size is further increased. For small ROI sizes, we observe an increase in false positives due to multiple detections of nearby objects as shown in figure \ref{fig:doubledet}. 

\subsubsection{Effect of secondary detector's input size on recall and computational overhead}
Table \ref{tab:hup_input} shows the performance and average GFLOPs per frame for different input sizes of the secondary detector tiny-yolov3 while the input size of the primary detector was fixed to 416. Based on our previous study, we fix the ROI size to 240 for this experiment. The recall is highest for the image size of 300 and we observe a decrease in the recall if we increase the image size beyond 300. Increasing the input size for the secondary detector with small input ROIs will increase the upscaling of the ROI before it is processed by the object detector and lead to an increase in noisy detections. The memory footprint of the network increases as the input size is increased. Based on the recall numbers and GFLOPs for the network, we choose an image size of 300 for the secondary detector. For all our further experiments, we fix the input image size of the secondary detector to 300 and the ROI size to 240. 
\begin{table}[t!]
    \centering
    {
	\captionsetup{justification=centering}
	\caption{Resource and recall comparison for different input sizes of the secondary detector }
	\setlength{\belowcaptionskip}{-10pt}
	\begin{tabular}{c c c }
		\hline
		\textbf{Image size} & \textbf{Recall} & \textbf{GFLOPs} \\
		\hline
		200 & 0.53 & 131.4  \\
		 256& 0.53 & 171.4  \\
		 300& 0.56 & 211.4  \\ 
		 350& 0.51& 266.4  \\

		416 & 0.52 & 346.4  \\
		 \hline
	\end{tabular}
	\label{tab:hup_input}
	}
\end{table}

\subsection{Evaluation}

\subsubsection{Experiments with different detectors}
\begin{table}
{
\caption{Performance and computational overhead for different fusion systems}
\centering
	\begin{tabular}{p{0.7cm} p{1.35cm} p{1.35cm} p{0.25cm} p{0.25cm} p{0.35cm} p{0.35cm} p{0.45cm}}
		\hline
		Mode& P det & Sec det. & Rcll$\uparrow$ &Prcn$\uparrow$ & FN$\downarrow$ & TE$\downarrow$ & GF$\downarrow$ \\
		\hline
		Fusion & yolov3-spp & tiny-yolov3 &  \textbf{0.51} & 0.68 & 1835&1.78 & 211.4\\
		Fusion & yolov3-spp& yolov3-spp &  \textbf{0.63} & 0.75 &  1398& 12.03 & 1791.4 \\
		Fusion & tiny-yolov3&  tiny-yolov3 &  \textbf{0.55} & 0.81 & 1833 & 1.50 & 150.6 \\
		Fusion & yolov3-spp & SSDlite & \textbf{0.51} & 0.67 & 1851&1.32 & 87.9\\
		Fusion & SSDlite & SSDlite & \textbf{0.48} & 0.69 & 1978 & 1.10  &22.3 \\

		\hline
	\end{tabular}
	{$P det$=Primary detector, $Sec det$=Secondary detector, $Rcll$=Recall, $Prcn$=Precision,$FN$=False Negatives, $TE$=Avg. Energy(J) per frame, $GF$= avg GFLOPs/frame }
	\label{tab:eval_many}
	}
	
\end{table}

\begin{table}
{
\caption{Comparison of our fusion system with baselines}
\centering
	\begin{tabular}{p{0.8cm} p{2.1cm} p{1cm} p{0.25cm} p{0.25cm} p{0.35cm} p{0.35cm} p{0.35cm}}
		\hline
		Mode& P det.(im sz) & Sec det. & Rcll$\uparrow$ &Prcn$\uparrow$ & FN$\downarrow$ & TE$\downarrow$ & GF$\downarrow$ \\
		\hline
		Fusion & yolov3-spp(416) & SSDlite & \textbf{0.51} & 0.67 & 1851&1.32 & \textbf{87.9}\\
		Fusion & SSDlite (416) & SSDlite & \textbf{0.48} & 0.69 & 1978 & 1.10  &\textbf{22.3} \\

			\hline
		\\
		\hline
		 & \textbf{Base Network} \\
		\hline 
		Base & yolov3-spp (416) &  & \textbf{0.42} & 0.92 & 2205&1.24 & \textbf{66.4}\\
		Base & yolov3-spp (1080) &  & \textbf{0.53} & 0.93 & 1769 & 2.48 &  \textbf{447.4}  \\
		\hline
		\\
		\hline
		 & \textbf{Prior Work} \\
		\hline 
		Fusion &   BIRANet-FFPN &  &\textbf{0.53} & 0.73 & 1762 & 1.71 & \textbf{207} \\
		\hline
	\end{tabular}
		{$P det$=Primary detector, $Sec det$=Secondary detector, $im sz$=Input image size for detector, $Rcll$=Recall, $Prcn$=Precision,$FN$=False Negatives, $TE$=Avg. Energy(J) per frame, $GF$= avg GFLOPs/frame }
	\label{tab:trackerEval_nuscenes}
	}
	
\end{table}


Table \ref{tab:eval_many} shows the performance of the fusion algorithm using different primary and secondary object detectors. The table shows the Average Recall (AR), Average Precision (AP), False Negatives, Total Energy consumption per frame and average GFLOPs per frame for the object detection task with an IoU of 0.4. The image sizes for the primary detector and the secondary detector are kept to 416 and 300 respectively with an ROI size of 240. According to Table \ref{tab:eval_many}, we get the highest recall for the system with yolov3-spp as both primary and secondary detector, however, it requires memory of 1791.4 GFLOPs per frame which is very high for a system with  limited computational resources. Fusion networks with tiny-yolov3 and SSDlite have similar performance in terms of average recall, however there is a big difference between their memory requirements. Network with yolov3 as primary and tiny-yolov3 as secondary detector requires more than twice the memory that the fusion system of yolov3 and SSDlite as primary and secondary detector respectively. Hence, we choose SSDlite as our secondary detector due to its low complexity and memory footprint. For SSDlite systems, yolov3 as the primary detector has the best recall however it requires nearly four times more computational power than the framework with SSDlite as the primary detector. In a system with very low computational power, a framework with SSDlite as both primary and secondary detector could be chosen over the yolov3-SSDlite framework with a slight decrease in detection accuracy. The framework with SSDlite as the primary detector will be referred to as 'SSDlite-SSDlite' while the framework with yolov3 as the primary detector will be 'yolov3-SSDlite.' 

\subsubsection{Comparison with baselines} In Table \ref{tab:trackerEval_nuscenes}, we compare the performance of our SSDlite fusion systems with two single yolov3-spp networks operating at input image  sizes of 416 and 1080. The yolov3 network with an input size of 1080 has a bigger receptive field for small objects and hence scores better recall than its counterpart with an input size of 416. Our both fusion systems (yolov3-SSDlite and SSDlite-SSDlite) outperform the baseline yolov3 with an input size of 416 with an improvement of 21\% and 14\% in the recall respectively. The single yolov3 detector with an input size of 1080 performs slightly better than our fusion systems however requires nearly 5 and 20 times  more resources than our yolov3-SSDlite and SSDlite-SSDlite fusion systems respectively. This demonstrates the high performance of our fusion systems while having very low memory requirements.

\begin{figure}
    \centering
    \includegraphics[width=\linewidth]{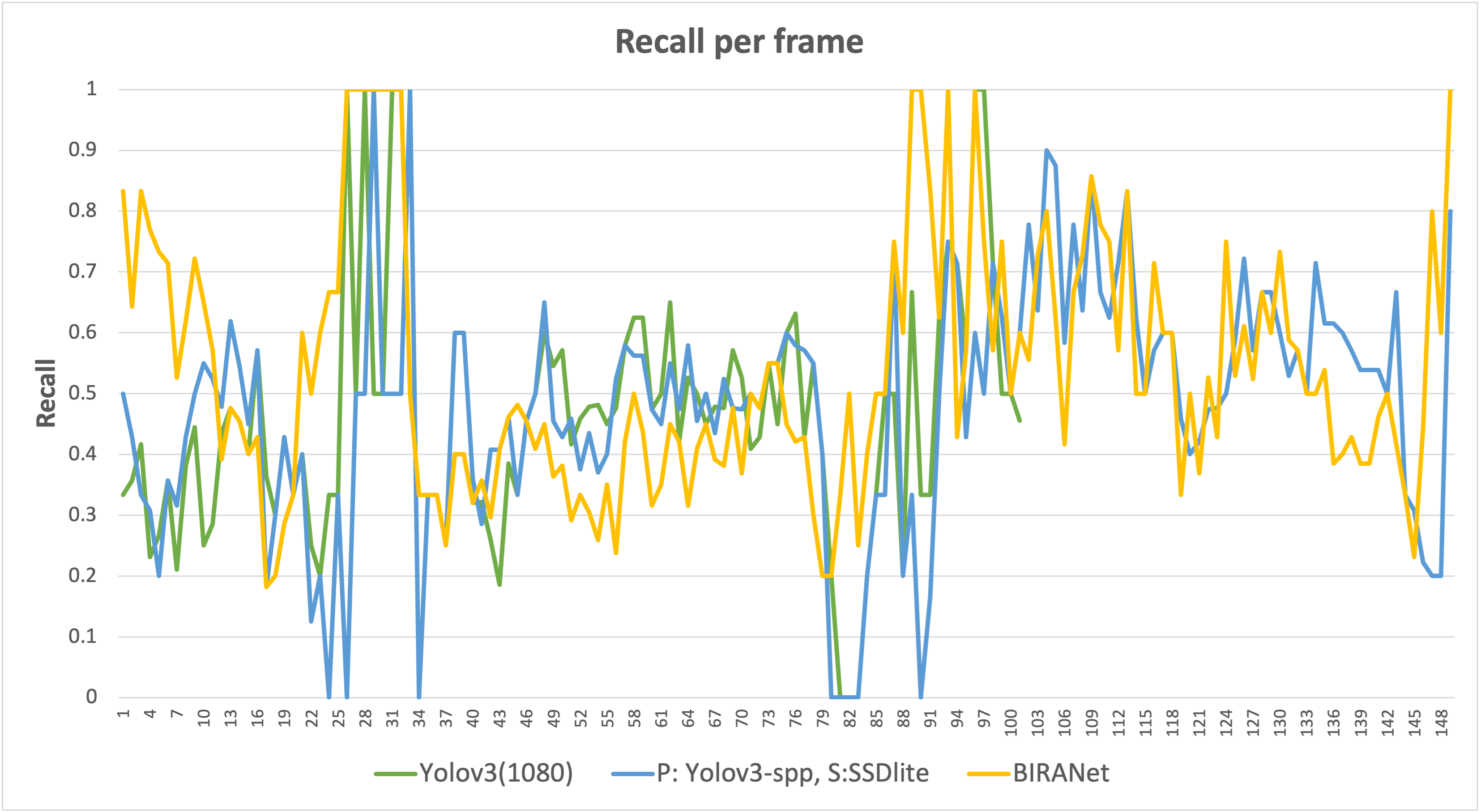}
    \caption{{Recall per frame for different detectors}}
    \label{fig:recall_analysis}
\end{figure}

We also evaluated BIRANet\cite{bira} on our dataset and observed only 4\% and 10\%  increase in recall compared to our SSDlite-SSDlite and Yolov3-SSDlite fusion systems respectively even though it uses high performing Faster R-CNN with a feature pyramid network for detection. The memory requirements for BIRANet are also very high compared to our fusion systems, requiring almost 9 times computational memory compared to our SSDlite-SSDlite fusion system. We also plot per frame recall for BIRANet, baseline yolov3 with 1080 input size and yolov3-SSDlite fusion system in figure \ref{fig:recall_analysis}. We observe that our fusion system has very a similar recall as compared to other analyzed networks over the sequence.  Thus our system is most suitable for deployment in real world resource efficient systems with the promise of better performance at low memory consumption.

\begin{figure}
    \centering
    \includegraphics[width=\linewidth]{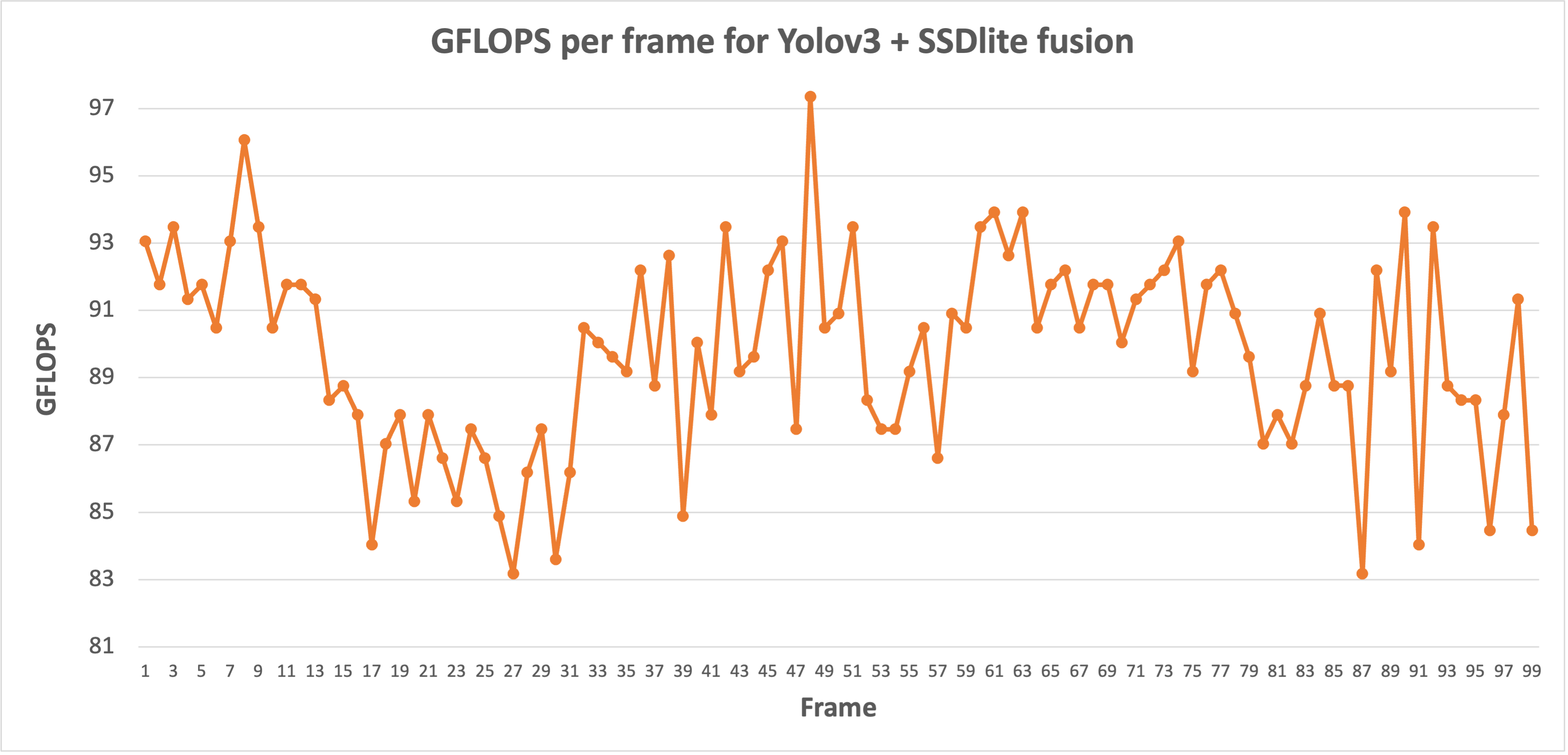}
    \caption{{GFLOPs per frame for fusion system with yolov3 and SSDlite as primary and secondary detector respectively}}
    \label{fig:energy_analysis}
\end{figure}

Figure \ref{fig:energy_analysis} shows the variation of memory requirements with the RGB frames for our fusion system with yolov3 and SSDlite. Unlike the tiling methods where a fixed number of proposals are created, our fusion system's number of proposals depend on detection made by radars. Hence the computational requirements change dynamically depending on the radar detections. This behavior is captured in figure \ref{fig:energy_analysis}. This ensures that the energy consumption is kept low for the dynamic scenes with very few objects in the frame. 





\begin{figure}
    \centering
    \includegraphics[width=\linewidth]{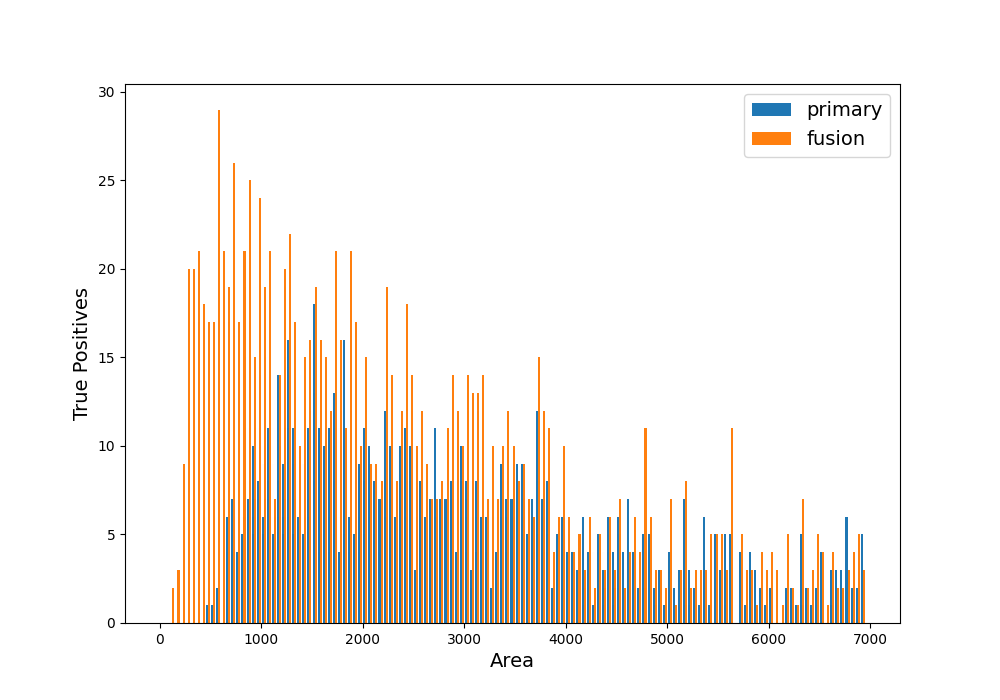}
    \caption{{Detection performance with respect to object area }}
    \label{fig:size_study}
\end{figure}

The effect of object size on the detection performance of our system is analysed in figure \ref{fig:size_study}. The y-axis in the figure represents the number of true positives and the x-axis represents the area of the detected true positive objects. We use SSDlite as the secondary detector and yolov3-spp as the primary detector with other parameters as mentioned above. 
In general, the performance of the fusion system and baseline are comparable for large objects. However, we see a significant increase in true positives with an area less than $1000$ $pixel$x$pixel$ for our fusion system as compared to the baseline primary detector. 

\begin{figure*}
     \centering
     \begin{subfigure}[b]{0.49\textwidth}
         \centering
         \includegraphics[width=\linewidth]{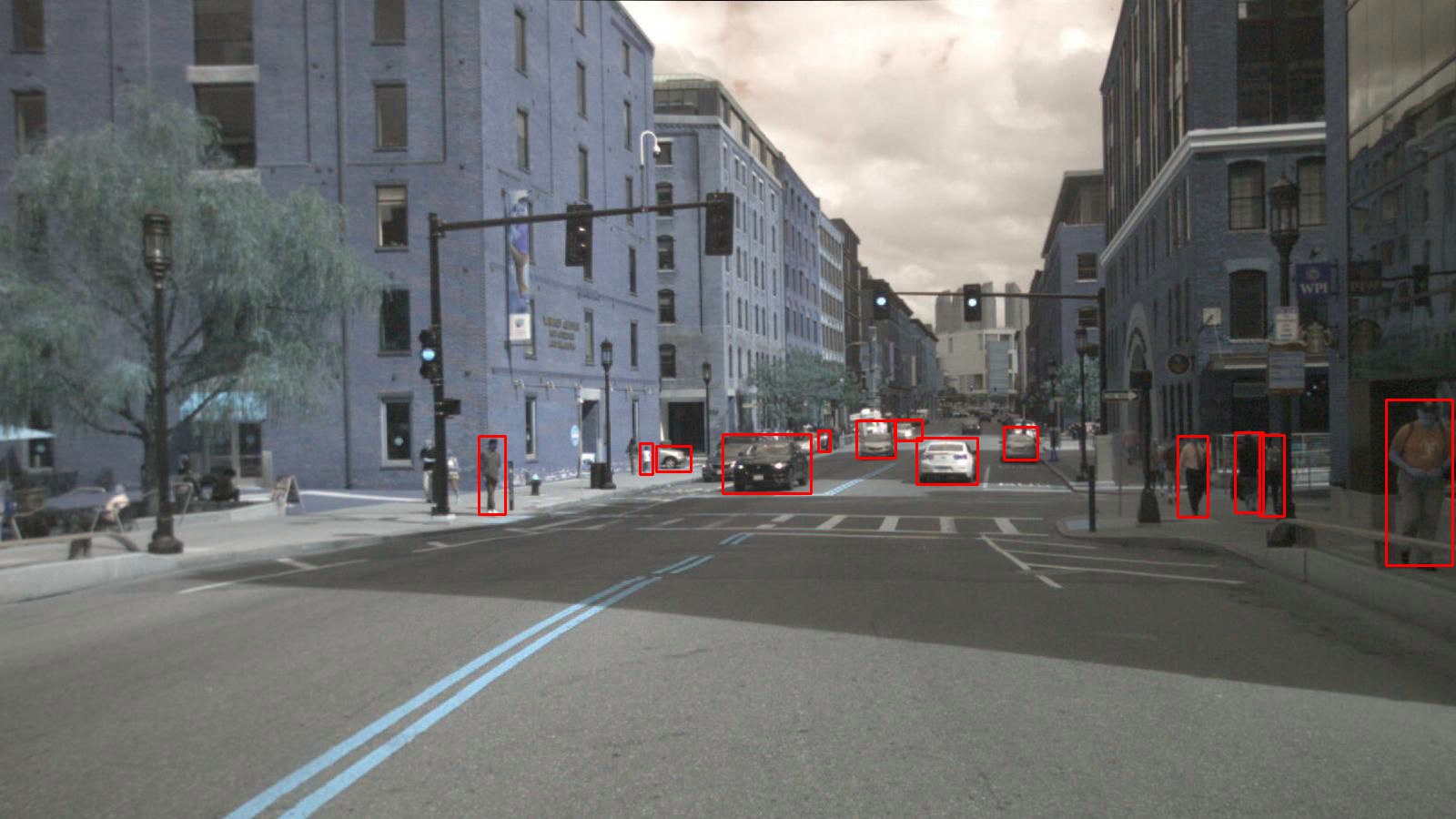}
         \label{fig:y equals x}
     \end{subfigure}
     \hfill
     \begin{subfigure}[b]{0.49\textwidth}
         \centering
         \includegraphics[width=\linewidth]{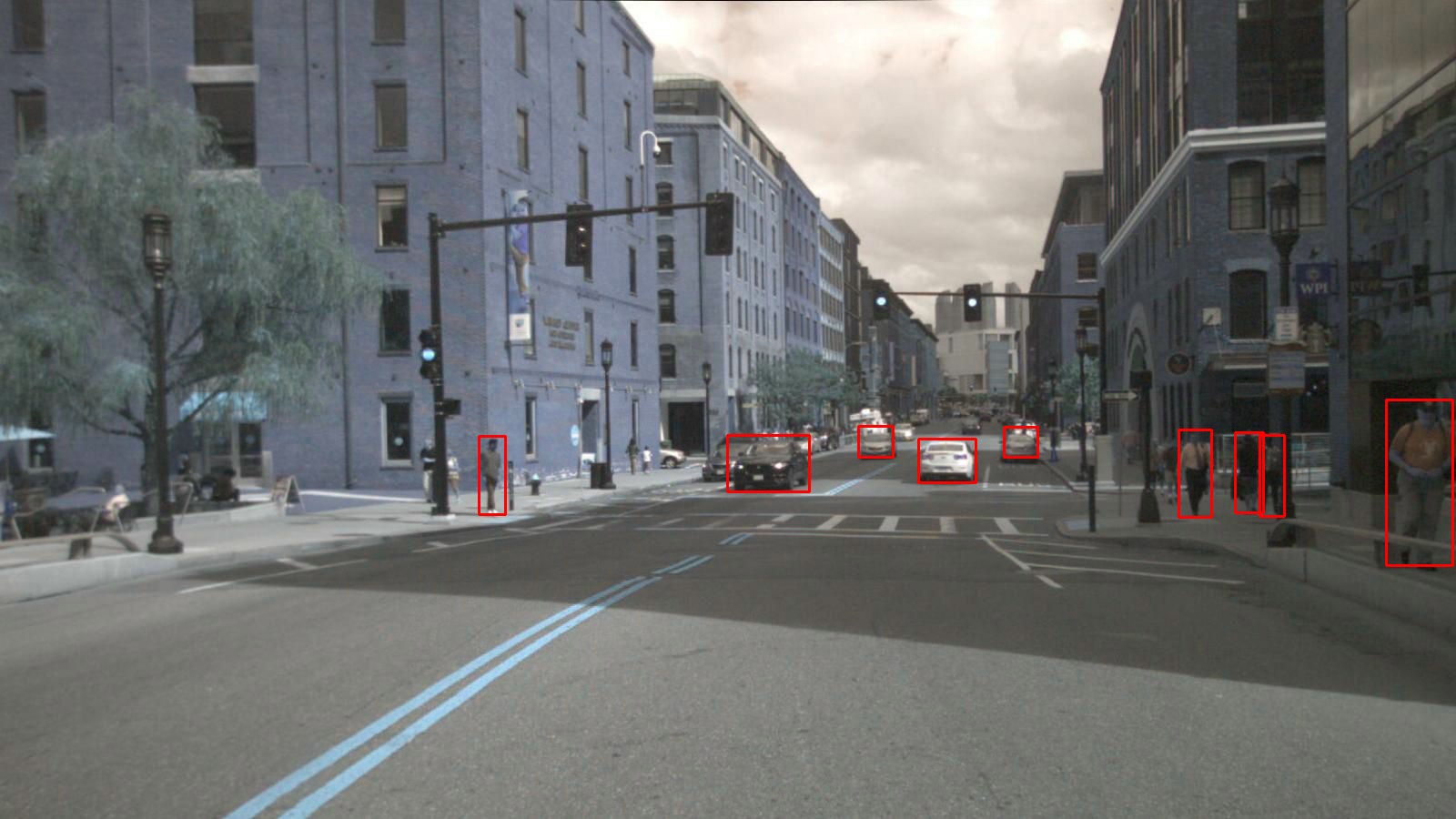}
           \label{fig:three sin x}
     \end{subfigure}
     \hfill 
     \begin{subfigure}[b]{0.49\textwidth}
         \centering
         \includegraphics[width=\linewidth]{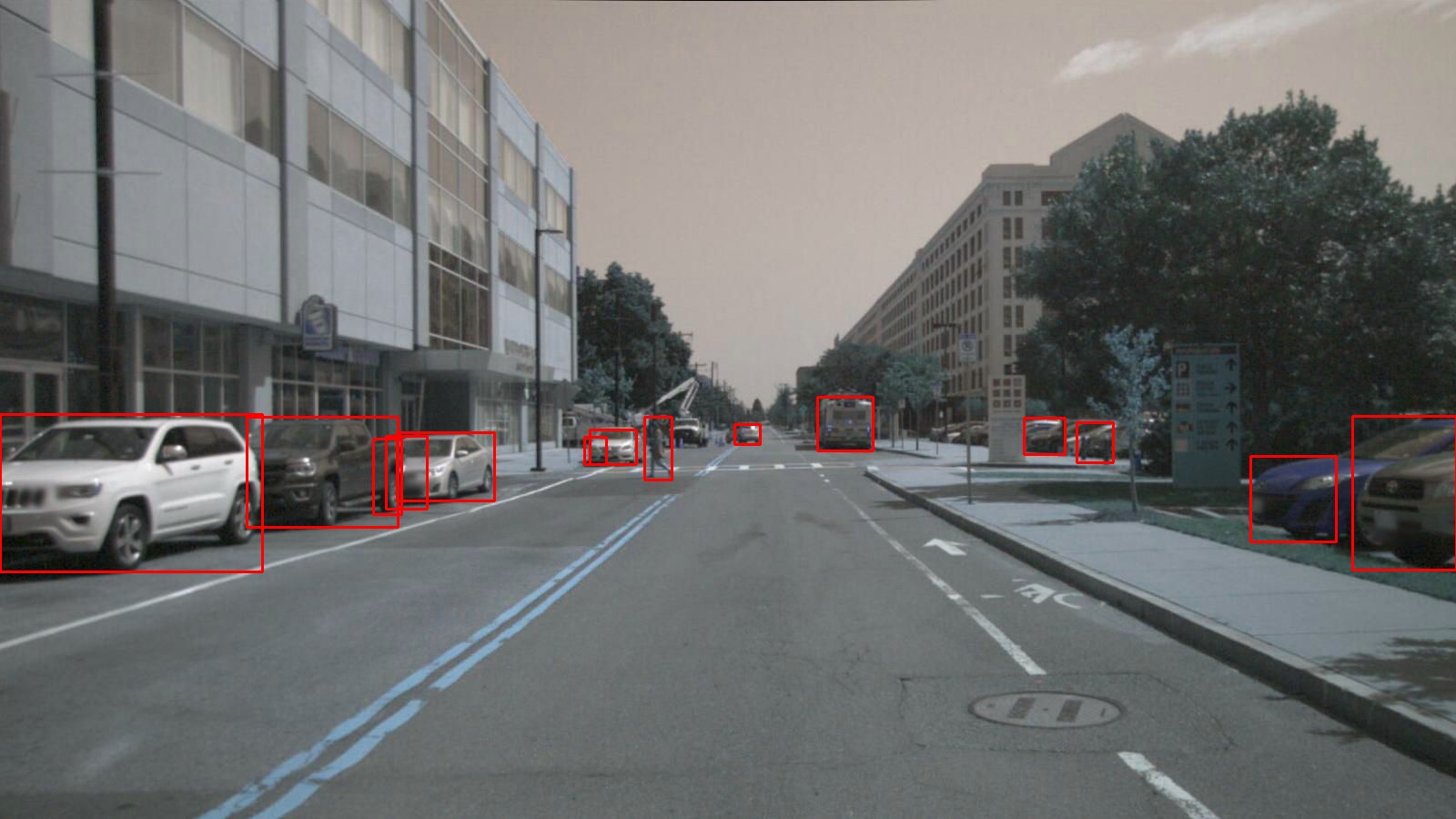}
         \label{fig:five over x}
     \end{subfigure}
     \hfill
     \begin{subfigure}[b]{0.49\textwidth}
         \centering
         \includegraphics[width=\linewidth]{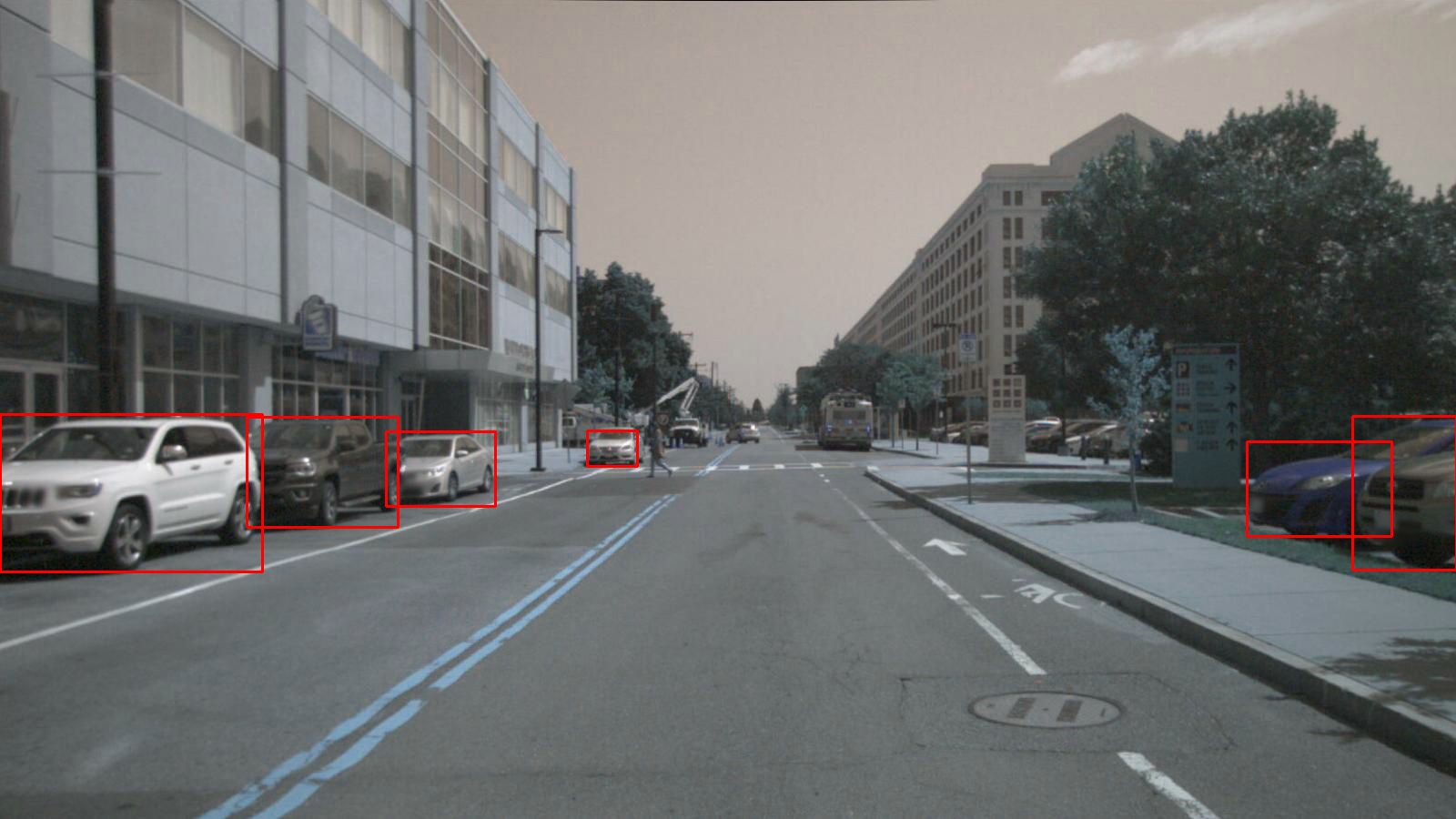}
         \label{fig:five over x}
     \end{subfigure}
     \hfill
     \begin{subfigure}[b]{0.49\textwidth}
         \centering
         \includegraphics[width=\linewidth]{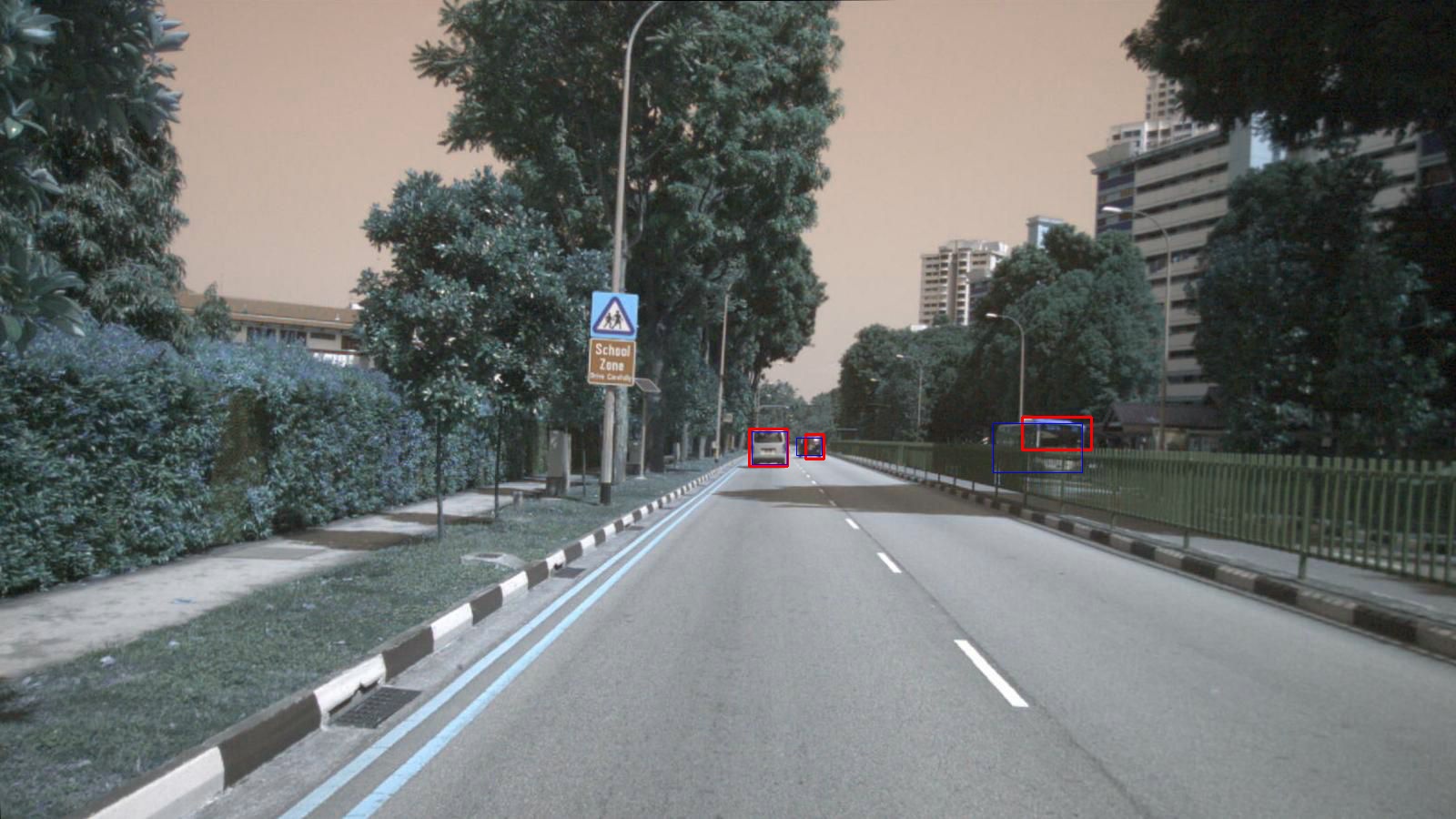}
         \label{fig:five over x}
     \end{subfigure}
     \hfill
     \begin{subfigure}[b]{0.49\textwidth}
         \centering
         \includegraphics[width=\linewidth]{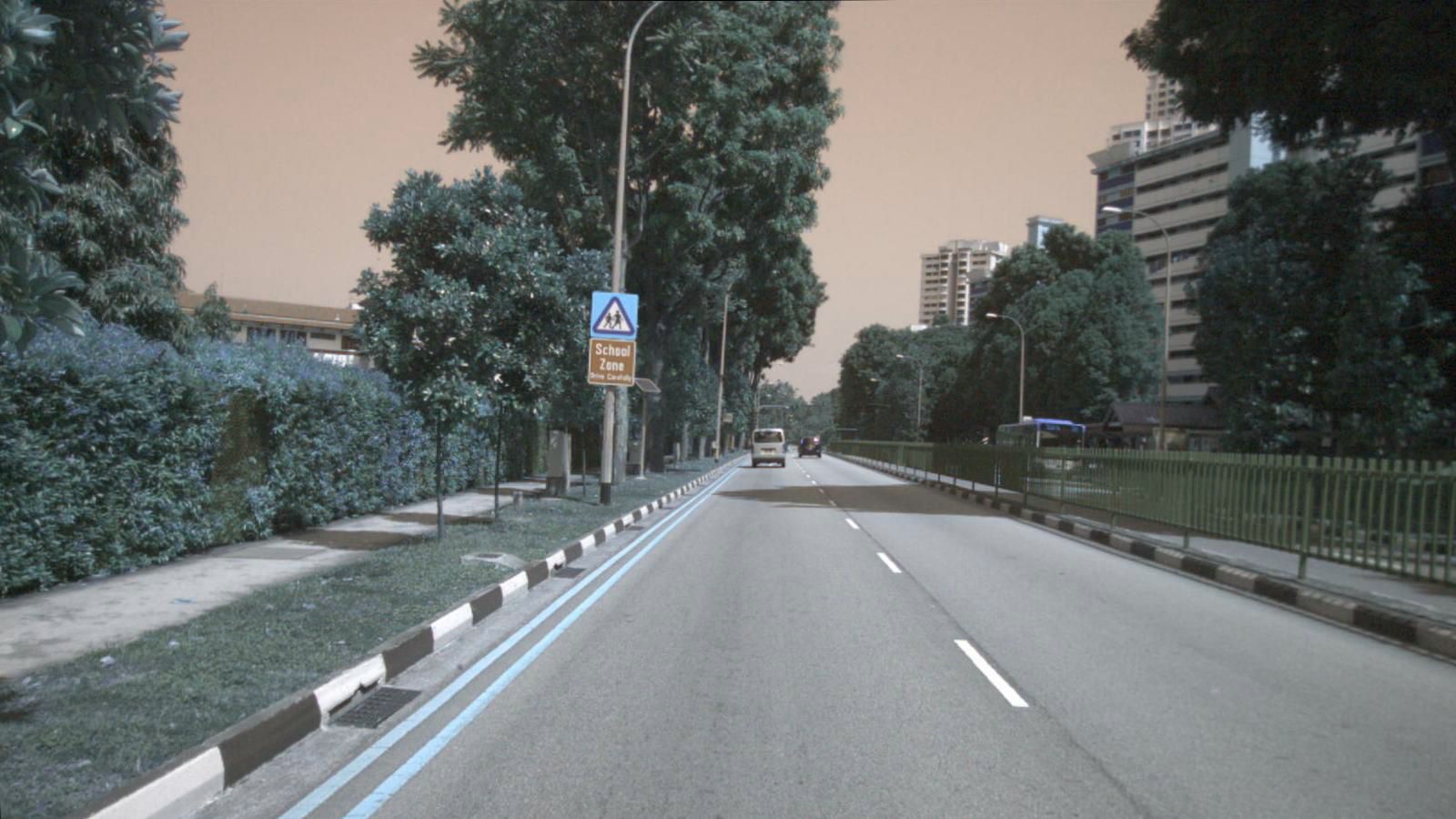}
         \label{fig:five over x}
     \end{subfigure}
     \hfill
     \begin{subfigure}[b]{0.49\textwidth}
         \centering
         \includegraphics[width=\linewidth]{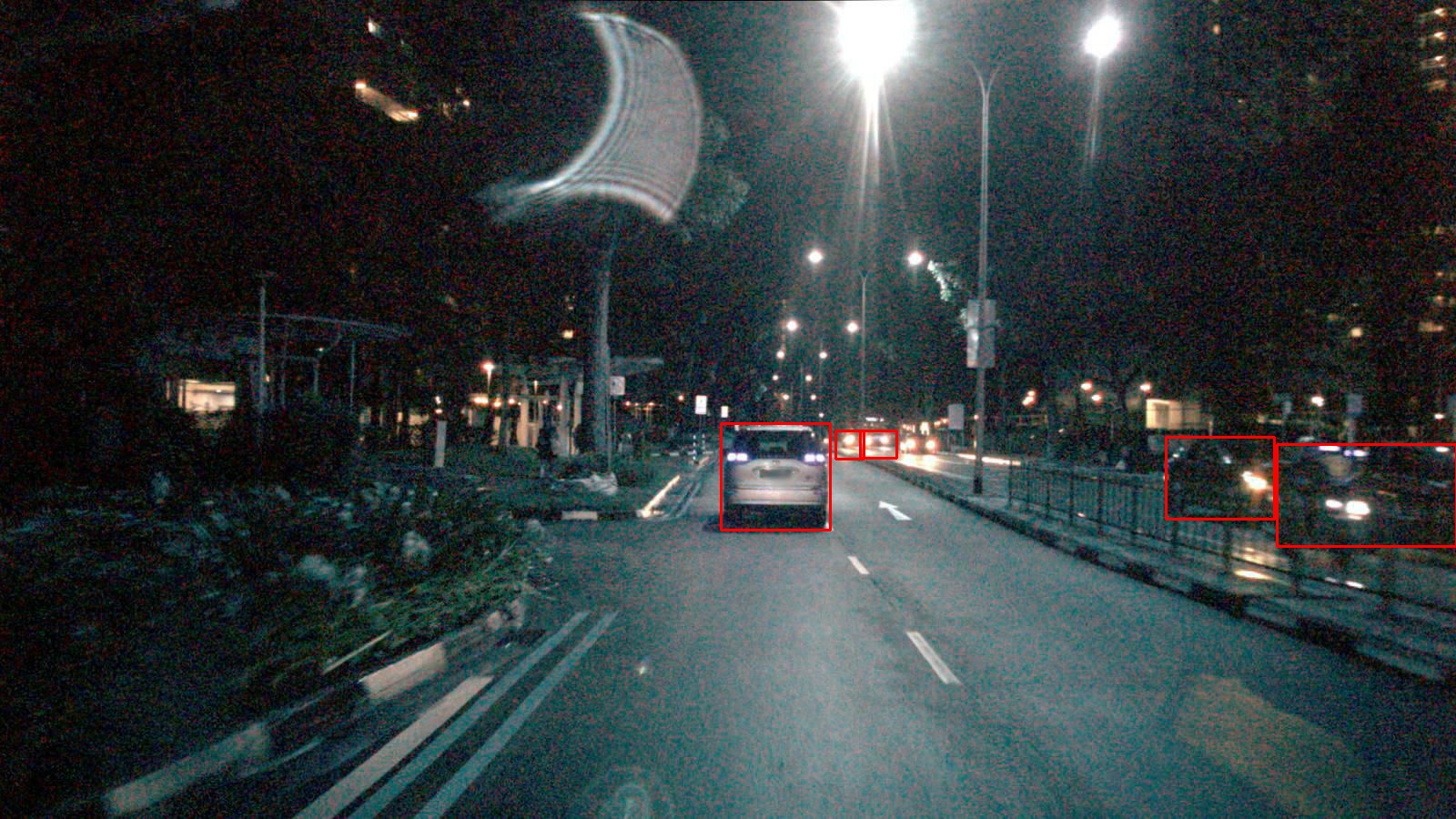}
         \caption{Detections by Fusion method}
         \label{fig:a_q}
     \end{subfigure}
     \hfill
     \begin{subfigure}[b]{0.49\textwidth}
         \centering
         \includegraphics[width=\linewidth]{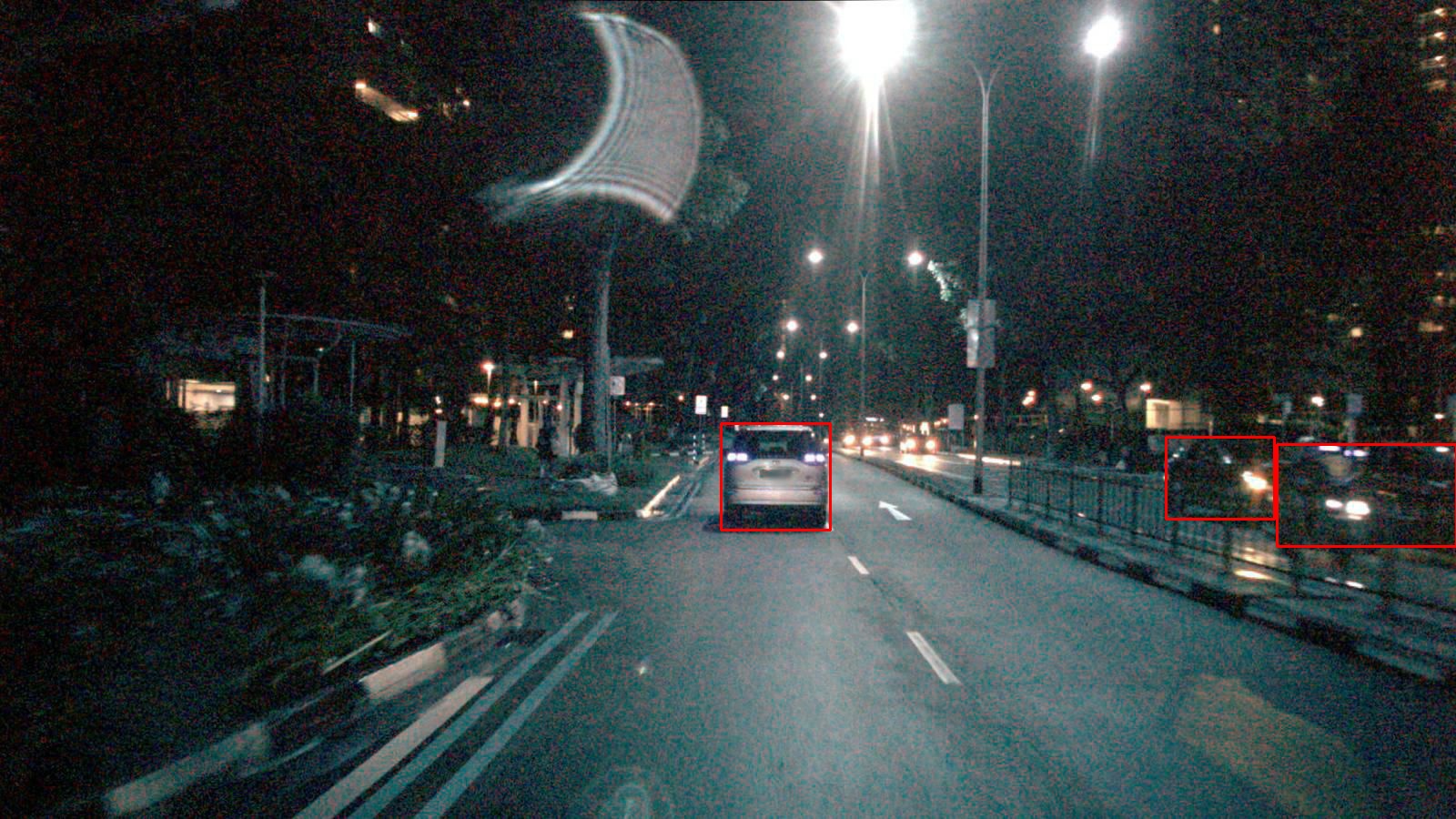}
         \caption{Detections by Primary Detector}
         \label{fig:d}
     \end{subfigure}
     \hfill
        \caption{Qualitative results of our fusion algorithm }
        \label{fig:qual results}
\end{figure*}

Figure \ref{fig:qual results} shows the qualitative results of our fusion system with yolov3 as the primary detector and SSDlite as the secondary detector. The results  support that our method performs very good in detecting small objects as well as the objects in low lightening or contrast regions. The image in the fourth row belongs to a dark sequence in nuScenes and our detectors have never seen night time sequences during training making object detection harder. 

\section{Conclusion}
Deep convolutional network based object detectors often fail in detecting small objects when operating at a fixed size input.  The lack of distinctive semantic features for the object and the mixing of background features when passed through the convolution network makes the small object detection problem very challenging. In this work, we have shown how radar could be used to create object proposals which when passed through a lightweight detector could help in the detection of small objects missed by the RGB detector. The proposed architecture uses a primary and a lightweight detector with low computational overhead to detect objects. The radar pointcloud is used to generate object proposal regions and these proposals are fed into the secondary detector to generate secondary detections. Detections generated with the primary detector with the full image as input are fused with the secondary detections via NMS. 
Our proposed fusion algorithm has the merits of high performance at longer distances, low computational requirements, high reliability and robustness. Experiments on  our annotated dataset show that our proposed SSDlite-SSDlite fusion method outperforms baseline primary yolov3 detector with a 14\%  increase in recall while having a very small computational overhead of  22.3 GFLOPs as opposed to the baseline's 66.4  GFLOPS.

\bibliographystyle{unsrt}
\bibliography{ref}

\end{document}